 \documentclass[final,authoryear,3p,times,twocolumn]{elsarticle}

\usepackage{amssymb}
\usepackage{amsmath}
\usepackage{graphicx}
\usepackage{color}
\usepackage{wasysym}
\usepackage[]{algorithm2e}

\newcommand{\bq}{\begin{equation}}
\newcommand{\eq}{\end{equation}}

\DeclareMathOperator*{\argmax}{arg\,max}

\newsavebox{\ieeealgbox}

\def\x{{\mathbf x}}

\usepackage{graphicx}

 \usepackage{lineno}

\journal{arXiv.org}

\begin{document}

\begin{frontmatter}



\title{Employing Weak Annotations for Medical Image Analysis Problems}


\author[doc]{Martin~Rajchl\corref{cor1}} \ead{m.rajchl@imperial.ac.uk}
\author[doc]{Lisa M. Koch}
\author[doc]{Christian Ledig}
\author[doc]{Jonathan Passerat-Palmbach}
\author[aic]{Kazunari Misawa}
\author[nag]{Kensaku Mori}
\author[doc]{Daniel Rueckert}

\cortext[cor1]{Corresponding author.}

\address[doc]{Dept. of Computing, Imperial College London, UK }
\address[aic]{Aichi Cancer Center, Nagoya, JP}
\address[nag]{Dept. of Media Science, Nagoya University, JP }

\begin{abstract}
To efficiently establish training databases for machine learning methods, collaborative and crowdsourcing platforms have been investigated to collectively tackle the annotation effort. However, when this concept is ported to the medical imaging domain, reading expertise will have a direct impact on the annotation accuracy. In this study, we examine the impact of expertise and the amount of available annotations on the accuracy outcome of a liver segmentation problem in an abdominal computed tomography (CT) image database. In controlled experiments, we study this impact for different types of weak annotations. 
To address the decrease in accuracy associated with lower expertise, we propose a method for outlier correction making use of a \emph{weakly labelled atlas}. Using this approach, we demonstrate that weak annotations subject to high error rates can achieve a similarly high accuracy as state-of-the-art multi-atlas segmentation approaches relying on a large amount of expert manual segmentations. Annotations of this nature can realistically be obtained from a non-expert crowd and can potentially enable crowdsourcing of weak annotation tasks for medical image analysis.
\end{abstract}


\end{frontmatter}
%
%
\setlength{\parindent}{0cm}

\section{Introduction}
%
%
In the recent past, collaborative and crowdsourcing platforms  \citep{estelles2012towards} have been investigated for their ability to obtain large amounts of user interactions for the annotation of image databases. Particularly, the capacity to outsource simple human intelligence tasks to a crowd population and simultaneously draw from client computing resources for interfacing, are being increasingly appreciated in the imaging community \citep{mckenna2012strategies,maier2014can,mavandadi2012biogames}. First studies employing collaborative \citep{haehn2014design,rajchl2016learning} or crowd sourcing platforms \citep{maier2014can,albarqouni2016aggnet} via web interfaces have been proposed for biomedical image segmentation. Since such interfaces often have limited capacity to interact with image data, weak forms of annotations (\emph{e.g.} bounding boxes, user scribbles, image-level tags, \emph{etc.}) have been investigated to reduce the required annotation effort. Recent studies have shown that placing bounding box annotations is approximately 15 times faster than creating pixel-wise manual segmentations \citep{lin2014microsoft,papandreou2015weakly}.

However, in contrast to annotating natural images \citep{lin2014microsoft,russell2008labelme} or recognising instruments in a surgical video sequence \citep{maier2014can}, the correct interpretation of medical images requires specialised training and experience \citep{nodine2000nature,gurari2015collect}, and therefore might pose a challenge for non-expert annotators, leading to incorrectly annotated data \citep{cheplygina2016early}. Nonetheless, considering the limited resources of available clinical experts and the rapid increase in information of medical imaging data (\emph{e.g.} through high-resolution, whole-body imaging, \emph{etc.}) alternative approaches are sought. Particular challenges arise, when trying to develop machine learning based approaches that can be scaled to very large datasets (\emph{i.e.} population studies). Many of the currently available approaches require large amounts of labelled training data to deal with the variability of anatomy and potential pathologies.

\subsection{Related Work}
To reduce the annotation effort, many well-known studies propose segmentation methods employing simple forms of user annotations to obtain voxel-wise segmentations \citep{boykov2000interactive,rother2004grabcut,rajchl2017deepcut,koch2017multi}. While adjusting hyperparameters can be considered an interaction, in this study we concentrate on simplified forms of pictorial input \citep{olabarriaga2001interaction}, called weak annotations (WA). Such annotations have been extensively used in the literature, particularly in the context of medical object segmentation problems. \cite{boykov2000interactive} used user-provided scribbles (SC) or brush strokes as input and hard constraints to an interactive graphical segmentation problem. Similarly, \cite{baxter2015optimization}, \cite{baxter2017directed} and \cite{rajchl2012fast} expressed this problem in a spatially continuous setting by using prior region ordering constraints and exploiting parallelism via GPU computing. The GrabCut algorithm \citep{rother2004grabcut} employs rectangular regions (RR) as bounding boxes to both compute a colour appearance model and spatially constrain the search for an object. These spatial constraints further allow to reduce the computational effort \citep{pitiot2004expert}. The segmentation platform ITK-SNAP\footnote{\emph{http://www.itksnap.org/}} \citep{yushkevich2006user} combines RR with SC and employs a pre-segmentation (PS) to initialise an active contour model. 

While the above object segmentation methods concentrate on how to accurately compute segmentations based on WA, recent studies have examined how to efficiently acquire the required annotations. Collaborative annotation platforms such as LabelMe\footnote{\emph{http://labelme.csail.mit.edu/}} \citep{russell2008labelme} or \citep{lin2014microsoft} were proposed to distribute the effort of placing image annotations to a crowd of users. Such crowdsourcing approaches have been successfully used in proof-reading  connectomic maps \citep{haehn2014design}, identification of surgical tools in laparoscopic videos \citep{maier2014can}, polyps from computed tomography (CT) colonography images \citep{mckenna2012strategies} or the identification of the fetal brain \citep{rajchl2016learning}.
However, most studies concentrate on tasks that require little expertise of the crowd, as the objects to identify are either known from everyday life \citep{russell2008labelme,lin2014microsoft} or foreign to background context \citep{maier2014can}. \cite{russell2008labelme} and \cite{lin2014microsoft} concentrated on object recognition tasks in natural images, the latter constrained to objects "easily recognizable by a 4 year old". \cite{maier2014can} asked users to identify a foreign surgical object in a video scene and \cite{haehn2014design} provided an automated pre-segmentation to be corrected by users. \cite{mckenna2012strategies} compensated for the lack of expertise in reading the colonography images by improving image rendering, implementing a training module and using a large number of redundant users (\emph{i.e.} 20 knowledge workers per task). 
Expertise in the interpretation of medical images is largely acquired through massive amounts of case-reading experience \citep{nodine2000nature} and it has been shown that novices performed with lower accuracy than an average expert in tasks such as screening mammograms for breast cancer \citep{nodine1999experience,nodine2000nature}. In contrast to \emph{diagnostic interpretation} of medical images, automated segmentation pipelines merely require the \emph{identification} of anatomical structures (\emph{i.e.} it requires less expertise to identify the liver in a CT image than a lesion in the liver).   

\subsection{Contributions}
In this study, we examine types of commonly employed WA and investigate the impact of reducing the annotation frequency (\emph{i.e.} only annotating every $k$-th slice in a volume) and the expertise on the segmentation accuracy. For this purpose, we employ a well-known graphical segmentation method and provide weak image annotations as initialisation and constraint to the segmentation problem at hand.
We address the problem of liver segmentation from a database of abdominal CT images and corresponding \emph{non-redundant annotations}. It is of great importance for both planning for laparoscopic surgery and computed-assisted diagnosis \citep{wolz2012multi} and requires extensive manual annotation effort because of its high spatial resolution and large field of view. 
Further, we propose and evaluate how a \emph{weakly labelled atlas} can be used for the detection and removal of incorrect annotations and achieve similar accuracy using weak annotations as a state-of-the-art fully-supervised automated segmentation method \citep{wolz2012multi}. 

\section{Methods} 
To study their impact on accuracy, we simulate user annotations from expert manual segmentations $M$, subject to different expertise levels and annotation frequency:

\subsubsection*{Expertise} 
We assume that the task of placing an annotation itself is defined well enough to be handled by a pool of general users with any experience level. However, the correct identification of anatomical structures might pose a challenge to non-expert users. We define expertise as the rate of correctly identifying an object of interest in a 2D image slice extracted from a 3D volume. If an error occurs, the user annotates the wrong object or rates that the object is not visible in this slice. We define the error rate (ERR $\in [0,1]$), i.e. the frequency of misidentification, as a measure of expertise.
An annotation error is simulated by computing an annotation from another organ (\emph{e.g.} the kidney, instead of the liver) or by setting the slice to background (\emph{i.e.} the organ is not visible in this slice). 

\subsubsection*{Annotation Rate} 
We collect an equal amount of annotations from each of the three slice directions $d \in D$ at an annotation rate (AR $\in [0,1]$). When computing the AR, we annotate every $k$-th slice, where $k = AR^{-1}$.  Note, that each slice was annotated at most \emph{once}, \emph{i.e.} annotations are \emph{non-redundant}. 

\begin{figure*}[!h]
\centering
\includegraphics[width=0.95\linewidth]{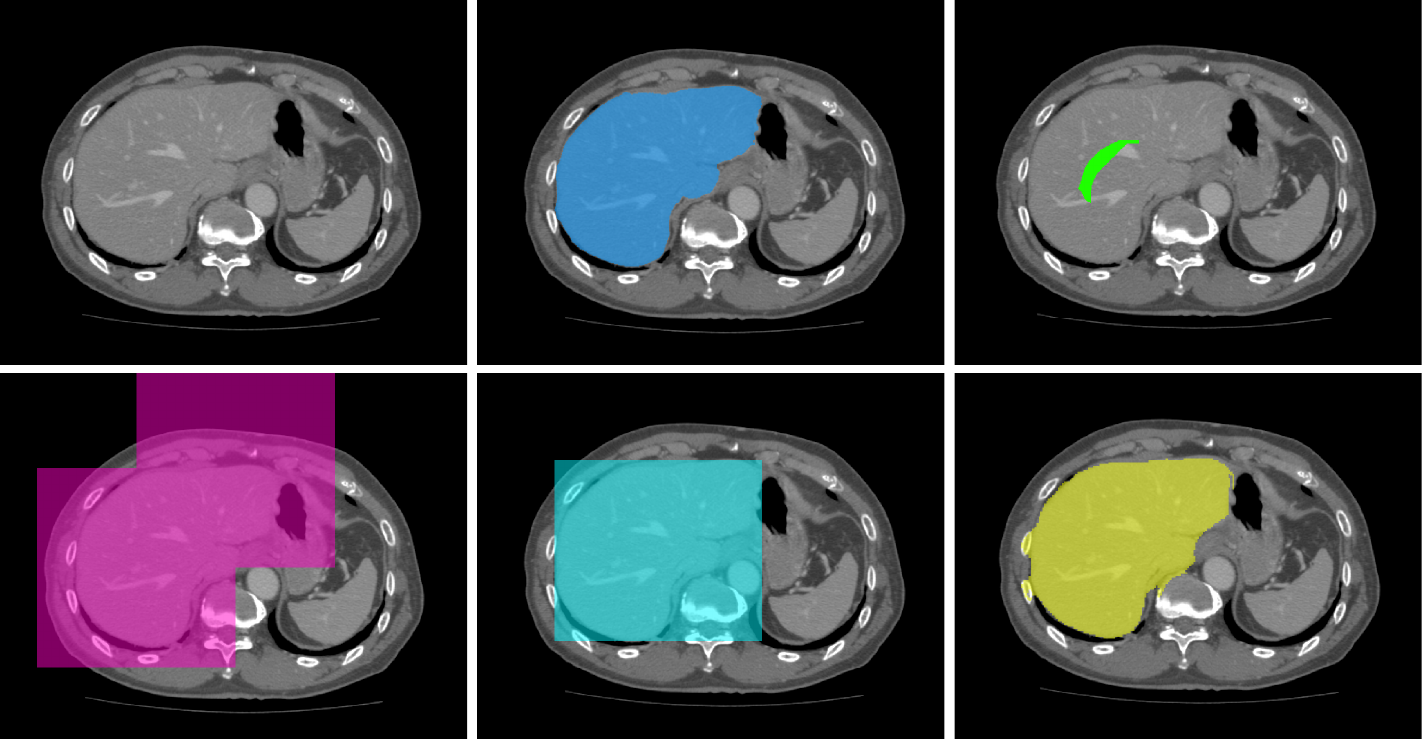}
\caption{Weak annotation types (top left to bottom right): image, expert manual segmentation (blue), scribbles (SC, green), binary decision making (BD, magenta), rectangular bounding box regions (RR, cyan) and merging of pre-segmentations (PS, yellow). }
\label{fig:ann_types}
\end{figure*}


\subsection{Annotation strategies}
For all experiments, we simulate following forms of weak annotations from expert manual segmentations $M$:

\subsubsection*{Brush strokes or scribbles (SC)}
Similarly to many interactive segmentation methods \citep{boykov2000interactive,rajchl2012fast,baxter2015optimization}, the user is asked to place a scribble or brush stroke into the object. We simulate placing scribble annotations by the iterative morphological erosion of the manual segmentation $M$ until a maximum of desired scribble size is reached. An example of such a generated SC label is depicted in Fig. \ref{fig:ann_types} (green).

\subsubsection*{Binary decision making (BD)}
The image is split into $N_{\textrm{BD}} = ds^2$ equally sized rectangular sub-regions, with the same number of splits ($ds$) per image dimension. For this type of weak annotation, a user is tasked to make a series of binary decisions on which sub-regions contain the object, such that all of the object is contained. We compute these weak annotations (BD) such that $\forall \mbox{BD} \, \cap \, M \neq \emptyset$. Fig. \ref{fig:ann_types} shows a BD (magenta) generated from $M$ (blue) for $ds = 4$. 

\subsubsection*{Rectangular (bounding box) regions (RR)}
Similarly to \citep{rother2004grabcut}, the user is asked to draw a tight rectangular region around the object. We compute a bounding box based on the maximum extent of $M$ within the respective image slice. An example RR (cyan) computed from $M$ (blue) is shown in Fig. \ref{fig:ann_types}.

\subsubsection*{Merging pre-segmentations (PS)}
Inspired by recent work in \citep{haehn2014design}, a user merges regions computed from an automated pre-segmentation method. We use a multi-region max flow intensity segmentation with the Potts energy, according to \citep{yuan2010continuous}: 
\bq
E(u) = \sum\limits_{\forall L} \int\limits_{\Omega}(D_L(x)u_L(x)+ \alpha_{\textrm{Potts}} |\nabla u_L(x)|)dx \, , 
\label{eq:potts_energy}
\eq 
\bq
s.t. \, u_L(x) \geq 0 \mbox{ and } \sum\limits_{\forall L}u_L(x) = 1
\eq
to obtain piecewise constant regions. The data fidelity term for each label $L = 1,\ldots,N_L$ is defined as the intensity L1-distance 
\bq
D_L(x)=|I(x)-l_L| \, , 
\label{eq:potts_dt}
\eq
where $l_L$ denotes the $L$-th most frequent intensity according to the histogram of the image volume. For all experiments, we fix $N_L$ = 16. The GPU-accelerated solver was provided with the ASETS library \citep{rajchl2016hierarchical}. After convergence, a discrete label map is calculated voxel-wise as $\argmax_l u_L(x)$.

\begin{figure*}[h!]
\centering
\includegraphics[width=0.95\linewidth]{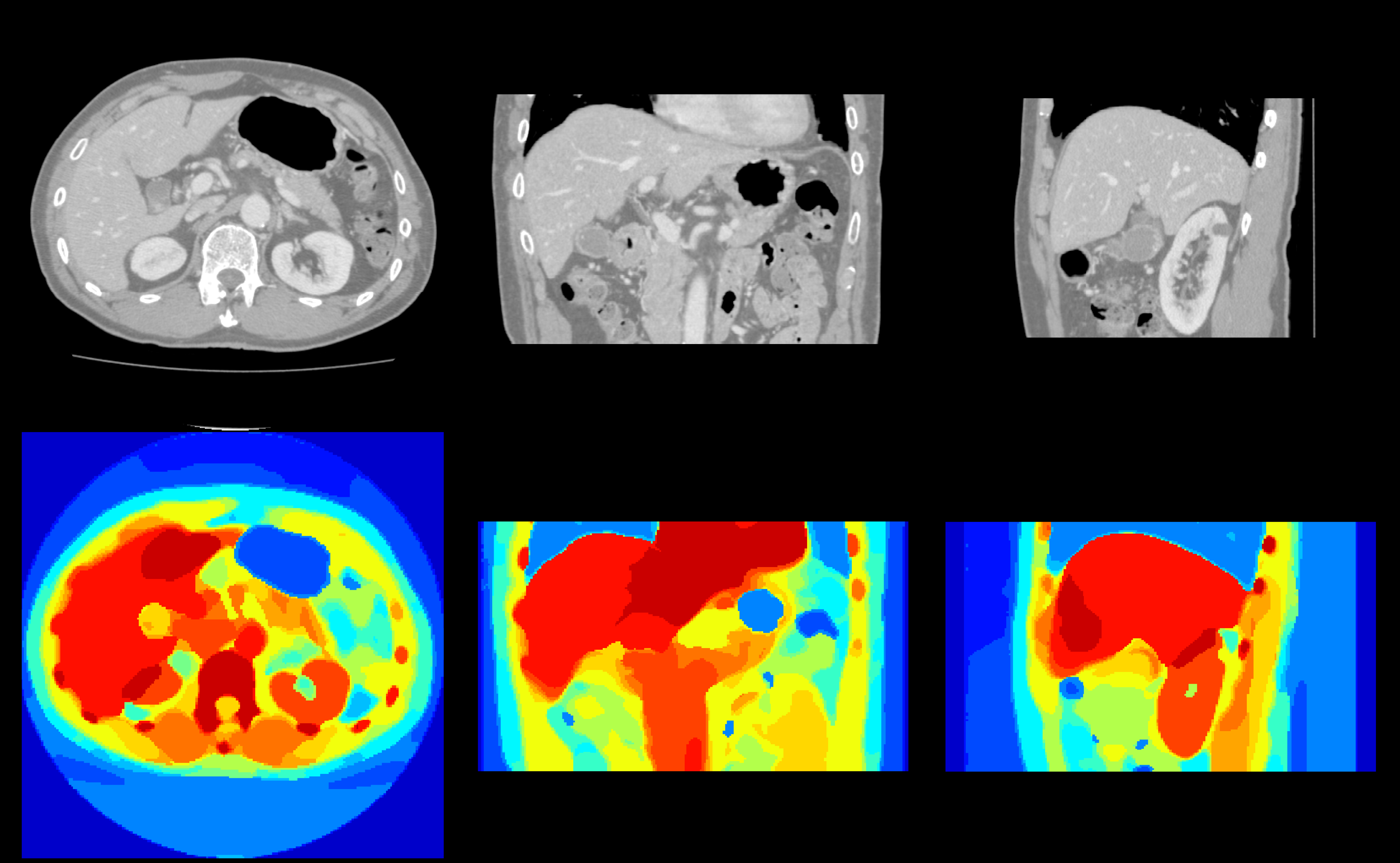}
\caption{Example pre-segmentation (PS) results on an abdominal CT volume: Top row (from left to right): CT slice images in transverse, coronal and sagittal direction. Bottom row: Corresponding PS segmentation labels after $\argmax_l u_L(x)$, when minimising \eqref{eq:potts_energy}, using \eqref{eq:potts_dt}, $N_L = 16$ and $\alpha_{Potts} = 0.05$.}
\label{fig:potts}
\end{figure*}

%

To obtain connected individual segments, the obtained segmentation is subsequently partitioned via 4-connected component analysis. Given such pre-segmentation (PS), the user is tasked to merge subregions, such that they contain the object of interest (\emph{i.e.} the liver). We simulate the merging similar to BD, such that $\forall \mbox{PS} \, \cap \, M \neq \emptyset$. A simulated PS annotation is shown in Fig. \ref{fig:ann_types} in yellow and the corresponding $M$ in blue. 

\subsection{Annotations as Segmentation Priors}
\label{sec:fuse_ann}
To be employed as priors in a volume segmentation problem, the annotations from individual slice directions $d \in D$ need to be consolidated to account for voxels $x$ located on intersecting slices:   \\
The binary SC annotations from all slice directions $d \in D$ are combined to a volume annotation $SC_{Vol}$
\bq
SC_{Vol}(x) \, = \, \cup_{d=1}^D \, SC_d(x) \, \, ,
\eq
and employed as foreground samples $S_{FG} = SC_{Vol}$. \\

All binary annotations $A_d \in \{BD, RR, PS\}$  and all unannotated slices $U_d$ are similarly combined to volumes to establish $S_{BG}$. Note, that $A_d(x) = 1$ denotes the user rated the location $x$ as "foreground" and $U_d(x) = 1$ denotes, that the user has not seen this location.
\bq
A_{Vol}(x) = \, \cup_{d=1}^D \, A_d(x) \,;  \, \, \, \, U_{Vol}(x) = \, \cap_{d=1}^D \, U_d(x) \,;
\eq
The background samples $S_{BG}$ are computed as all voxels $x$ that are outside $A_{Vol}$ and \emph{are} annotated:
\bq
S_{BG}(x) \, = \, \neg \, A_{Vol}(x) \, \, \cap  \,\,  \, \neg \, U_{Vol}(x) . 
\eq

The resulting samples $S_{FG}$ and $S_{BG}$ can then be used to compute intensity models or to enforce spatial constraints. For all experiments, we employ SC annotations as priors for foreground voxels and \{BD, RR, PS\} annotations as priors for background voxels. For each of these three combinations (\emph{e.g.} SC and BD, \emph{etc.}), we calculate $S_{FG}$ and $S_{BG}$ for each volume image to be segmented.



\subsection{Segmentation Problem}
The method employed to obtain a segmentation can be considered as a black box to be replaced by any specialised pipeline that suits a specific problem. For our experiments, we employ a well-known interactive flow maximisation \citep{boykov2000interactive,rajchl2012fast} approach to compute image segmentations from the input annotations $A$, subject to a certain $AR$ and $ERR$. For this purpose, we use the continuous max-flow solver \citep{yuan2010study,rajchl2016hierarchical} supporting GPU acceleration and allowing us to tackle the computational load for our experiments. We find a solution by minimising an energy $E(u)$ defined for the labelling or indicator function $u$ at each voxel location $x$ in the image $I$, $ \mbox{ s.t. } u(x) \in [ 0, 1 ]$ as, 

\bq
E(u) = \int\limits_{\Omega}(D_s(x)u(x) + D_t(x)(1-u(x))+ \alpha|\nabla u(x)|)dx  \, , \\
\label{eq:binary_e}
\eq
Here, the data fidelity terms $D_{s,t}(x)$ are defined as the negative log-likelihood of the probabilities $\omega_{1,2}$, computed from normalised intensity histograms of the foreground (FG) and background (BG) region, respectively,
\bq
D_s(x) \, = \, -log(\omega_1 (I(x))) \, \mbox{ and } \,  
D_t(x) \, = \, -log(\omega_2 (I(x))) \, , \\
\label{eq:ll_data_term}
\eq
as described in \citep{boykov2001interactive}. Additionally, we employ a soft spatial constraint by setting the cost for regions annotated as FG and BG, to a minimum:

\begin{align}
D_s(x) = 0,  \, \,  \forall x \in  \mbox{FG}; \, \, D_t(x) = 0,  \, \,  \forall x \in \mbox{BG};
\label{eq:soft_constraints}
\end{align}

Consolidated volume annotations (see Section \ref{sec:fuse_ann}) are used to compute samples $S_{FG}$ and $S_{BG}$ of FG and BG, respectively. $S_{FG}$ and $S_{BG}$ are subsequently employed to compute $\omega_{1,2}$ in \eqref{eq:ll_data_term} and as spatial constraints in \eqref{eq:soft_constraints}.
After optimisation of the energy in \eqref{eq:binary_e}, the resulting continuous labelling function $u$ is thresholded at 0.5 to obtain a discrete segmentation result for the FG, as described in \citep{yuan2010study}.

\subsection{Outlier Detection \& Removal}
\label{sec:outlier_detection}
We propose a method for quality assessment to mitigate the impact of annotation errors on the accuracy of the segmentation results (\emph{e.g.} when using databases labelled by crowds with low expertise). Note, that contrary to other studies \citep{mckenna2012strategies,lin2014microsoft}, we do \emph{not} require redundant annotations for outlier detection. Instead, we propose to make use of redundant information in the \emph{flawed} and \emph{weakly labelled} atlas database and retrieve similar image slices and their annotations in the fashion of multi-atlas segmentation pipelines \citep{wolz2012multi,aljabar2009multi}.

If spatial variability is accounted for (\emph{e.g.} through registration), we can retrieve  slices from other atlas volumes and use their annotations to compute an agreement measure to rate a given annotation. For this purpose, we borrow from the concept of the SIMPLE method \citep{langerak2010label}, where an iteratively refined agreement is used to assess the quality of individual atlases in a multi-atlas segmentation approach.

\subsubsection*{Weakly Labelled Atlas as Quality Reference}
We assume that $S$ subjects $s_i=\{s_1,\ldots,s_S\}$ have been weakly (and potentially erroneously) annotated and aim to automatically detect the slices of each subject $s_i$ that have an annotation of insufficient quality (\emph{e.g.} the wrong organ has been annotated or the organ was present, but not detected). 
In the following, the $j$-th slice of subject $s_i$ in direction $d$ is denoted by $v_{i}^{j,d}$. For each slice in the database $v_{i}^{j,d}$, we first find a subset of the most similar spatially corresponding images $v_{q}^{j,d}$ of the subjects $s_q$ in the \emph{weakly labelled atlas} using a global similarity measure, such as the sum of squared differences. 
We then calculate a consensus segmentation $\bar{O_1}$ from the annotations of these anatomically similar image slices using mean label fusion. For each of these selected atlas annotations, the overlap between the annotation and the estimated consensus segmentation is calculated with an accuracy metric.

For this purpose, we use the Dice similarity coefficient (DSC) between the regions $A$ and $B$ as a measure of overlap:
\bq
DSC = \frac{2 |A| \cap |B|}{|A| + |B|}
\label{eq:dsc}
\eq 

Using the mean regional overlap $\mu_\textrm{DSC}^1$ between the atlas annotations and the consensus segmentation $\bar{O_1}$, we can discard potentially inaccurately annotated atlas slices if their DSC with $\bar{O_1}$ is less than this average $\mu_\textrm{DSC}^1$. Following \citep{langerak2010label}, we calculate another fusion estimate $\bar{O_2}$ using the reduced subset of both anatomically similar and reasonably accurate annotations and calculate another mean DSC, $\mu_\textrm{DSC}^2$, and reject the annotations corresponding to $v_{i}^{j,d}$ if its DSC with $\bar{O_2}$ is less than $\mu_\textrm{DSC}^2$.

\begin{figure}[!ht]
\centering
\includegraphics[width=0.95\linewidth]{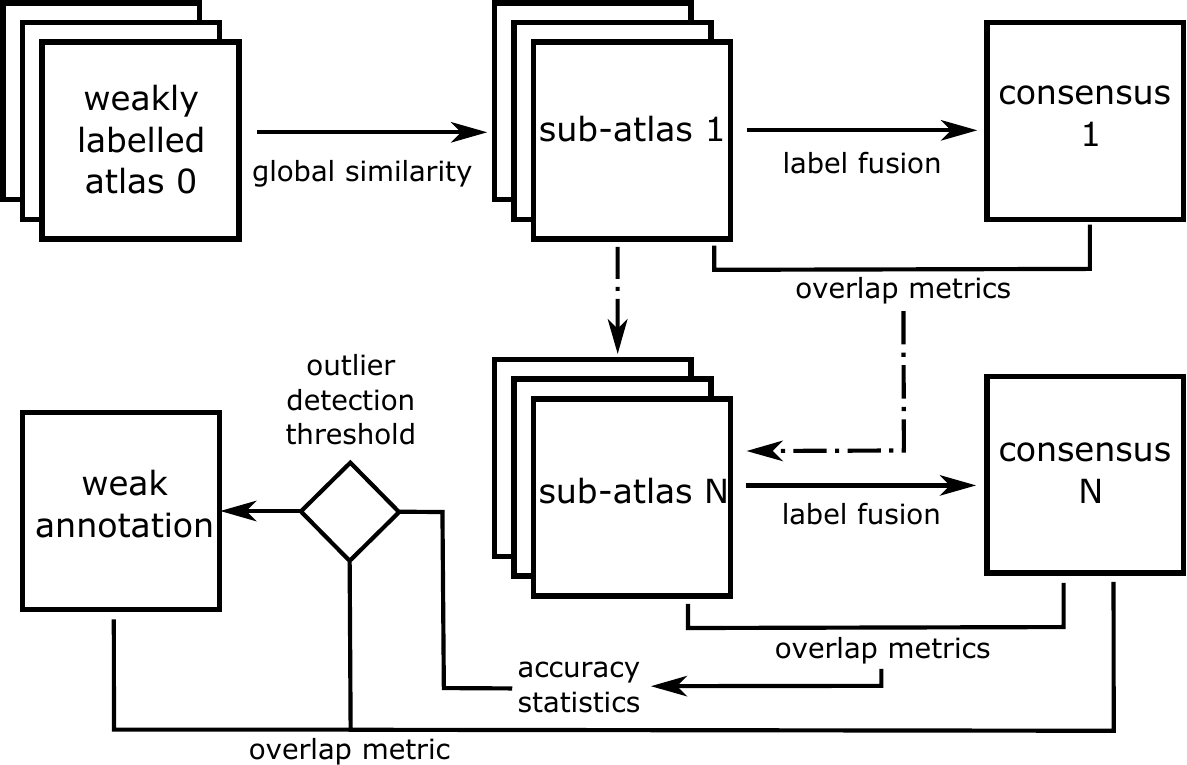}
\caption{Schematic illustration of the outlier detection and removal approach.}
\label{fig:outlier_detection}
\end{figure}

This procedure is repeated for each annotation in the database. Note that the \emph{weakly labelled atlas} can be built from the database itself so that no external/additional input is required. An illustration of the approach is provided in Fig. \ref{fig:outlier_detection}.

\begin{algorithm}[h]
 \KwData{weak annotation for $v_{i}^{j,d}$: $wa_{i}^{j,d}$\;
 corresponding WAs in the weakly labelled atlas $Q$: $wa_{q}^{j,d}$, $q \in Q$}
 \KwResult{$v_{i}^{j,d}$ is outlier: yes/no}
 $Q^1 \leftarrow \{ p \in Q: |Q^1|=N_\textrm{similar}$ and $\sum_{q\in Q^{1}}||v_{i}^{j,d}-v_{q}^{j,d}|| \rightarrow min$ \}\;
 i = 1\;
 \While{ i $\le$ N$_\textrm{iterations}$}{
  $\bar{O_i} \leftarrow$ MajorityVote$(wa_{q}^{j,d} ~ \forall q\in Q^{i}$)\;
  $\mu_i \leftarrow $ Average( Dice( $\bar{O_i}, wa_{q}^{j,d}$) ~ $\forall q\in Q^{i}$)\;
 $Q^{i+1} \leftarrow \{ p \in Q^{i}:$ Dice($\bar{O_i}, wa_{q}^{j,d}) \ge \mu_i\}$\;
$i \leftarrow i+1$
}
\eIf{Dice($wa_{i}^{j,d}, Q^{N_\textrm{iterations}}$) $\ge \mu_{N_\textrm{iterations}}$}
 {
   return yes\;
   }{
   return no\;
  }
\caption{Outlier detection using a weakly labelled, flawed atlas database.}
\end{algorithm}

\section{Experiments}
\subsubsection*{Image Database}
The image database used in the experimental setup consists of 150 (114 \mars, 36 \female) abdominal volume CT images with corresponding manual segmentations from expert raters. Available labelled anatomical regions include the liver, spleen, pancreas and the kidneys.  All scans were acquired at the Nagoya University Hospital with a TOSHIBA Aquilion 64 scanner and obtained under typical clinical protocols. The volume images were acquired at an in-plane resolution of 512 x 512 voxels (spacing  0.55 to 0.82 mm) and contain between 238 and 1061 slices (spacing 0.4 to 0.8 mm).


\subsubsection*{Pre-processing \& Generation of Weak Annotations}
Prior to the experiments, all volume image data were affinely registered using the NiftiReg library \citep{modat2010fast} (default parameters) to a random subject to spatially normalize the images and to account for variability in size. Weak annotations are generated for each slice in each direction $d$ in all volume images of the database, subject to the annotation rate $AR = \{1, 0.5, 0.33, 0.25, 0.1, 0.05, 0.01\}$ and the error rate $ERR = \{0, 0.05, 0.1, 0.25, 0.5\}$.

\subsubsection*{Liver Segmentation with Weak Annotations}
The weak annotations are fused to compute $S_{FG}$ and $S_{BG}$ (see \ref{sec:fuse_ann}) to subsequently compute the data terms $D_{s,t}(x)$ in \eqref{eq:ll_data_term} and the soft constraints in \eqref{eq:soft_constraints}. A continuous max-flow segmentation \citep{yuan2010study}, minimizing \eqref{eq:binary_e} is computed to obtain a segmentation result.
The regularisation parameter $\alpha = 4$ in \eqref{eq:binary_e} and the parameters $\alpha_{Potts} = 0.05$ and $N_L = 16$ in \eqref{eq:potts_energy} were determined heuristically on a single independent dataset. For the outlier detection, $N_{iterations}$ was set to 2. All experiments were performed on an Ubuntu 14.04 desktop machine with a Tesla 40c (NVIDIA Corp., Santa Clara, CA) with 12 GB of memory.

\subsubsection*{Experimental Setup}
20 consecutively acquired subject images are used as a subset to examine the impact of AR, ERR and type of weak annotation on the mean segmentation accuracy. An average DSC is reported as a measure of accuracy between the obtained segmentations and the expert segmentations $M$ for all the parameter combinations of ERR, AR and all examined annotation types (SC in combination with \{BD, RR, PS\}), resulting in 2100 single volume segmentations results.

Further, the proposed outlier detection (see Section \ref{sec:outlier_detection}) is employed using annotations from all 150 subjects. The annotations after outlier removal are used for repeated segmentations, resulting in additional 2100 segmentations. A study on atlas selection \citep{aljabar2009multi} suggests an optimal subset size for brain segmentation to be 20. We increased $N_{similar}$ of the globally similar atlases $Q^1$ to 30, to account for the variation in abdominal soft tissue organs, such as the liver.

A series of paired T-tests is computed to determine significant changes in accuracy at a $p = 0.05$ level between resulting DSC before and after outlier detection.

\section{Results}
\begin{figure}[!ht]
\centering
\includegraphics[width=0.95\linewidth]{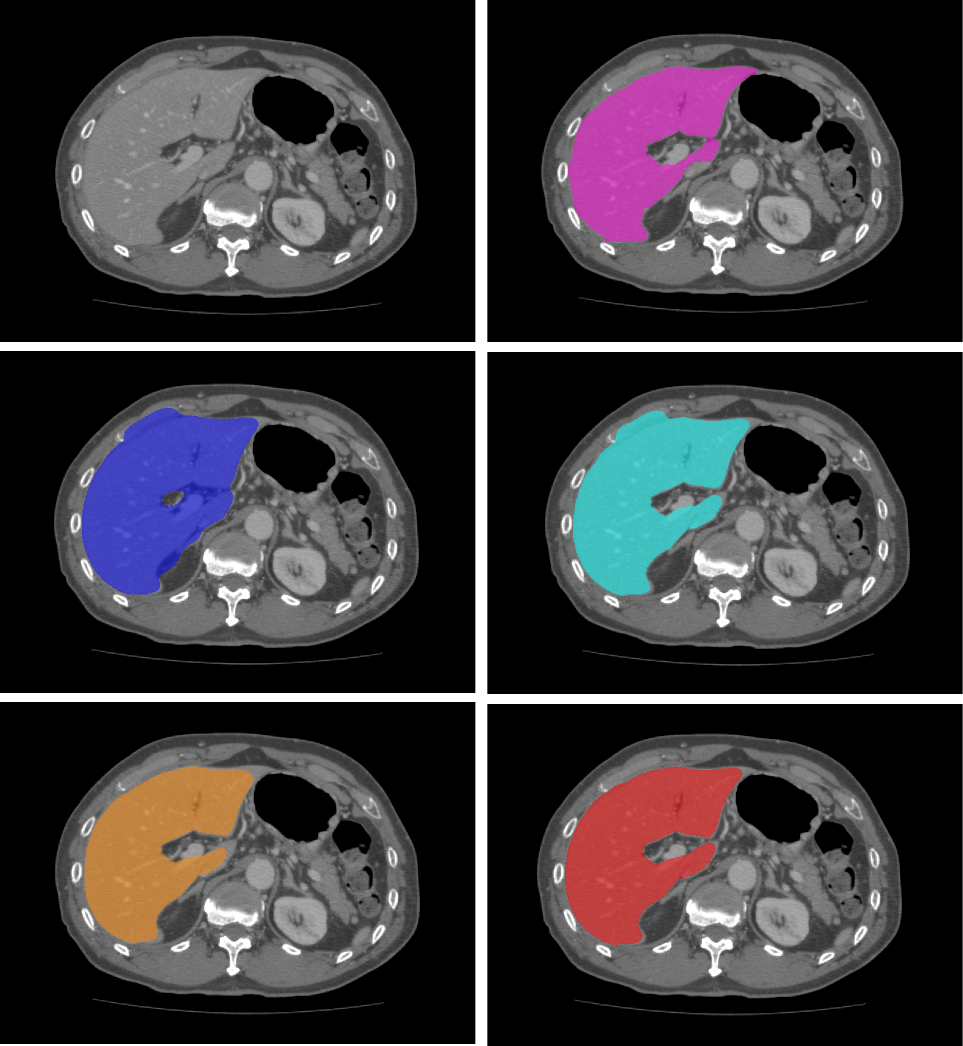}
\caption{Example segmentation results (from top left to bottom right): expert manual segmentation (magenta), segmentation with $< 0.8$ (blue), $\approx 0.85$ (cyan), $\approx 0.9$ (orange) and $\approx 0.95$ accuracy according to DSC. The colour coding is chosen to reflect those of accuracy matrices in Fig. ~\ref{fig:results_mean}.}
\label{fig:results_visual}
\end{figure}

\begin{figure}[!ht]
\centering
\includegraphics[width=0.95\linewidth]{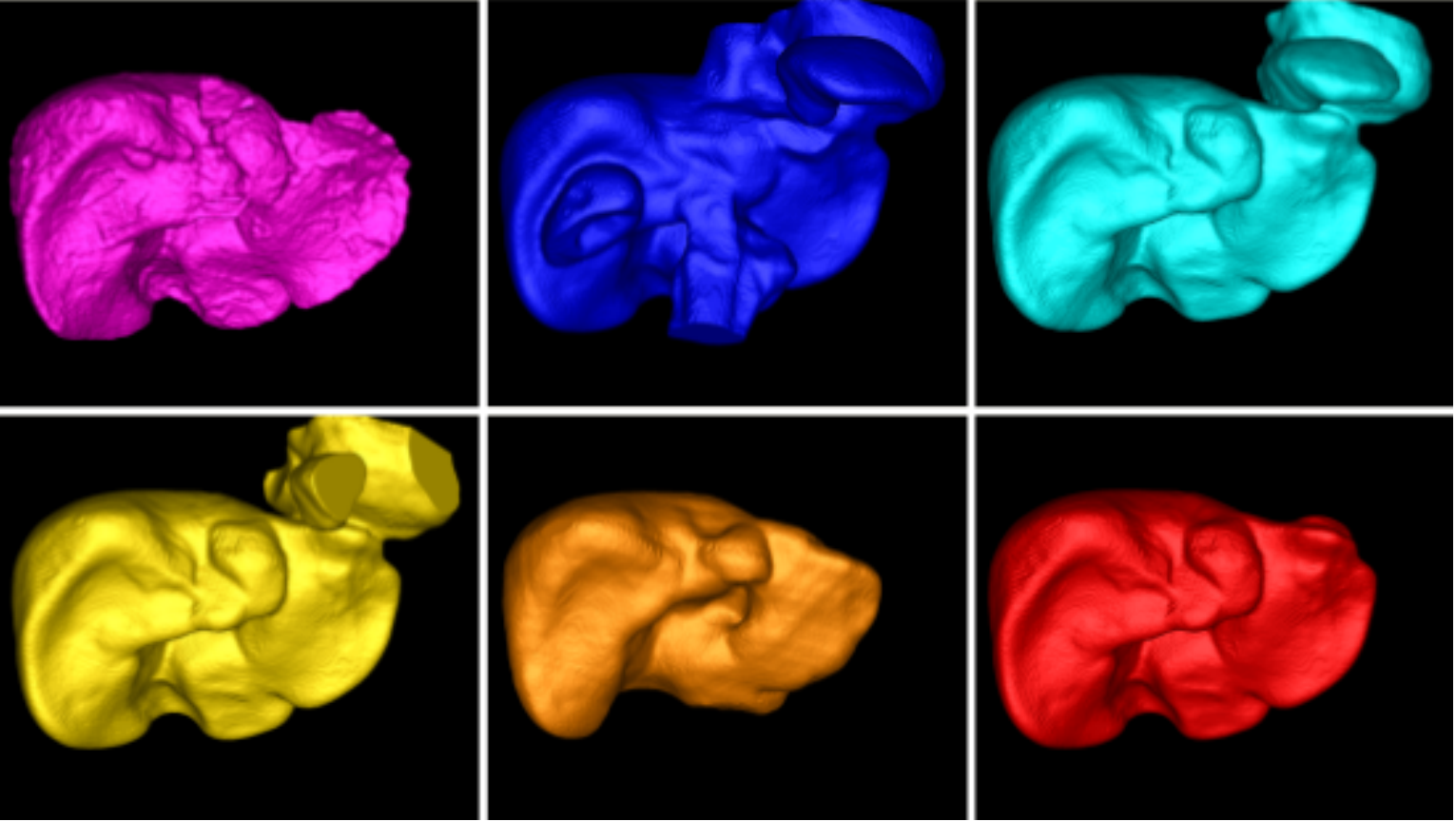}
\caption{Surface rendering of example segmentation results. The colour coding are chosen to reflect those of accuracy matrices in Fig. ~\ref{fig:results_mean}.}
\label{fig:results_visual_surf}
\end{figure}

\subsubsection*{Segmentation Accuracy}
Fig. \ref{fig:results_visual} depicts example segmentations of transverse slices on a single subject as comparative visual results with obtained DSC ranges. Visual inspection suggests that a DSC \textgreater 0.9 can be considered an acceptable segmentation result. A DSC of lower than 0.8 can be considered a segmentation failure. Mean accuracy results of all examined methods are shown in Fig. \ref{fig:results_mean}. 
The main contribution to a decrease in accuracy was observed to be high error rates. Without outlier correction, acceptable segmentation results could be obtained with all annotation types, down to an AR of 25\%. Using rectangular regions, this accuracy can still be obtained when annotating 1\% of available slices, when the ERR is simultaneously less than 5\%. In a densely annotated database (\emph{i.e.} AR = 100\%) more than 10\% of erroneous annotations lead to segmentation failure. This is particularly interesting for medical image analysis studies considering a crowdsourcing approach with \emph{non-redundant annotations} of non-experts. This effect still persists to a degree, after outlier correction at the highest tested ERR of 50\%.

\subsubsection*{Performance after Outlier Correction}
\begin{figure*}[!ht]
\centering
\includegraphics[width=0.8\linewidth]{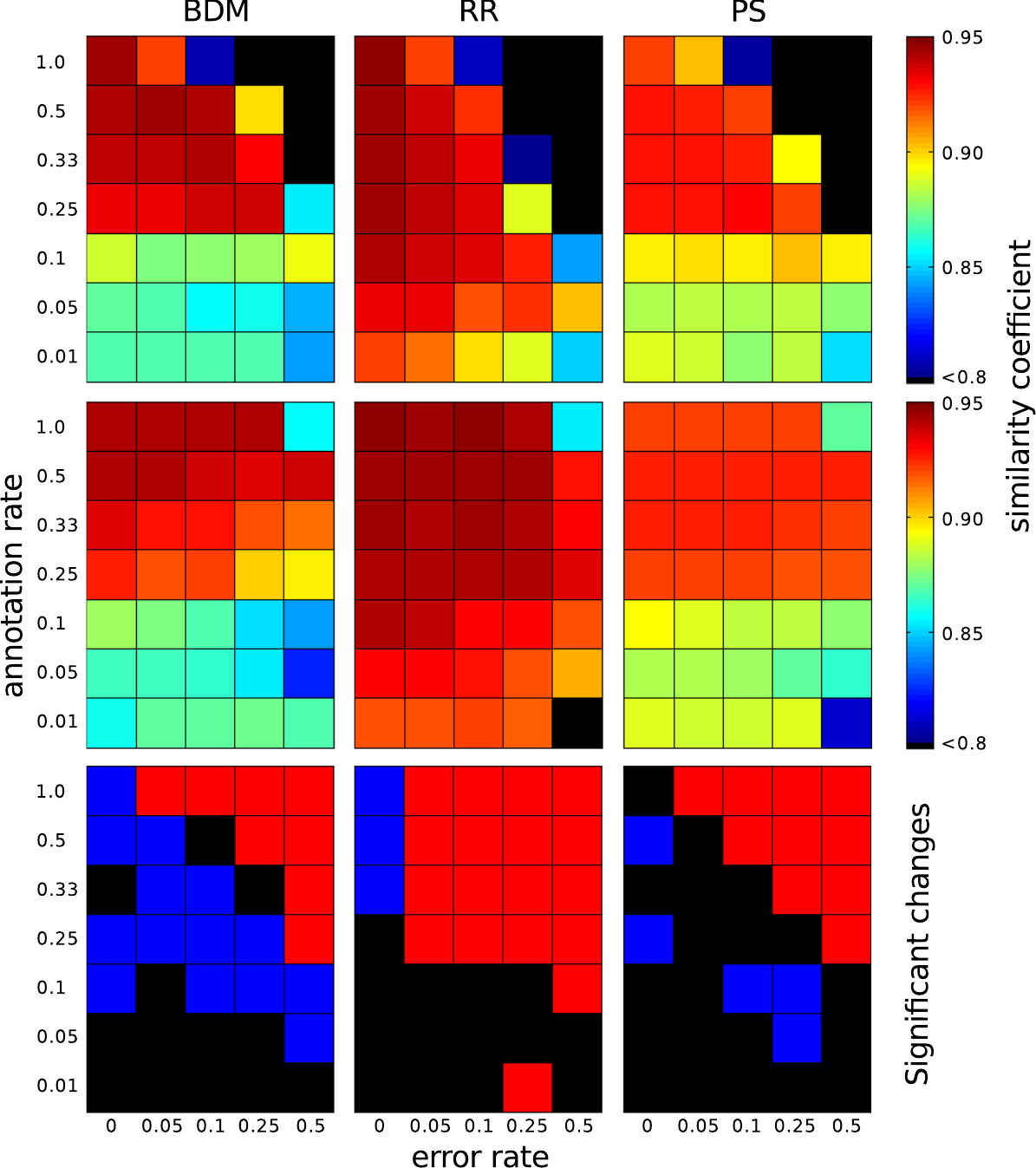}
\caption{Top Row: Accuracy results for each annotation type, subject to AR and ERR, without outlier detection. Middle Row: Accuracy results with proposed outlier detection. Bottom Row: Results of paired T-tests between top and middle row (p \textless 0.05). Increased and decreased mean accuracy are depicted in red and blue, respectively. Black elements show no significant difference.}
\label{fig:results_mean}
\end{figure*}

The mean accuracy improves after the proposed outlier correction, however slight decreases in accuracy are observed at lower error rates. This is mainly due to the decreased number of available annotations after correction. The differences in mean DSC after outlier removal range from  $-0.05$ to $+0.94$ (BD), $-0.0006$ to $+0.94$ (RR) and $-0.02$ to $+0.92$  (PS). Statistically significant changes are visualised in Figure \ref{fig:results_mean} (bottom row) and numerical ranges reported in Tab. \ref{tab:acc_diff}. 

\begin{table}[!h]
\label{tab:acc_diff}
\centering
\caption{Minimal and maximal changes in DSC accuracy after outlier removal, for all tested ERR and AR and each annotation type.}
\begin{tabular}{|l|c|c|c|}
\hline
ANN Type & \textbf{BD} & \textbf{RR} & \textbf{PS} \\
\hline
incr. $N$ (DSC) &  8 (+0.94) & 18 (+0.94) & 10 (+0.92) \\
decr. $N$ (DSC) &  14 (-0.05) & 3 (-0.0006) & 5 (-0.04) \\
\hline
\end{tabular}
\end{table}

\begin{table*}[!h]
\label{tab:acc_diff}
\centering
\caption{Mean segmentation accuracy with decreasing annotation rate (AR) and ERR = 0. All results as DSC [\%].}
\renewcommand{\arraystretch}{1.2}
\begin{tabular}{|l|c|c|c|c|c|c|c|}
\hline
& \multicolumn{7}{c|}{Annotation Rate (AR) [\%]}  \\ \hline
Type & 100 & 50 & 33 & 25 & 10 & 5 & 1 \\
\hline
BD & \, 94.3 (0.8) & \, 94.3 (1.2) & \, 94.0 (1.4) & \, 93.3 (2.2) & \, 88.7 (3.7) & \, 87.0 (4.0) & \, 86.9 (3.7) \\
RR & \, 94.6 (0.8) & \, 94.5 (0.8) & \, 94.4 (0.8) & \, 94.3 (0.8) & \, 94.1 (0.9) & \, 93.2 (1.2) & \, 92.1 (1.9) \\
PS & \, 92.1 (1.4) & \, 92.7 (1.4) & \, 92.9 (1.4) & \, 92.8 (1.8) & \, 89.6 (4.0) & \, 88.2 (4.8) & \, 88.8 (4.1) \\
\hline
\end{tabular}
\end{table*}

\section{Discussion}
In this study, we tested types of weak annotations to be used for liver segmentation from abdominal CTs. We examined the effects of different expertise levels and annotation rates of crowd populations for their impact on the segmentation accuracy outcome and proposed a method to remove potential incorrect annotations. In the conducted experiments each of the slices was annotated at most \emph{once}, reducing the effort associated with the acquisition of redundant annotations.

\subsubsection*{Segmentation Accuracy}
While the max-flow segmentation was employed without any problem-specific adaptions, it yields comparably high accuracy to a state-of-the-art hierarchical multi-atlas approach described in \citep{wolz2012multi}, where a mean DSC of 94.4\% was reported for the segmentation of the liver. Comparable accuracy was obtained when employing RR annotations at AR = 10\% and no errors present or for BD at AR = 50\% and ERR = 10\%. After the proposed outlier correction, using RR at an AR of 25\%, errors of up to 25\% yielded similarly high accuracy - a scenario realistic enough to be obtained by a non-expert crowd. Both BD and PS annotation types performed similarly robust to RR at higher annotation rates and yielded acceptable accuracy at AR down to 25\% and ERR of up to 25\% without outlier correction. 

\subsubsection*{Impact of Expertise and Annotation Rate}
An expected decrease in accuracy with both higher ERR and lower AR is observed for all annotation types. We report more robust behaviour of the RR annotations and at a wide range of ERR and at an AR of down to 5\%. For BD and PS annotations, AR of less than 25\% yielded insufficient accuracy, even at the highest expertise levels (\emph{i.e.} ERR = 0). For all annotation types, the presence of errors has a larger impact at high AR, suggesting that the total amount of incorrectly annotated image slices is related with segmentation failure, rather than its rate. Without correction, high ERR were tolerated in annotation rates of 1-10\%, however lead to segmentation failure (\emph{i.e.} DSC \textless 0.8) at higher AR. This suggests that an increased number of annotations is not beneficial if performed at an high error rate.   

\subsubsection*{Outlier Correction}
The proposed \emph{weakly labelled atlas}-based outlier detection approach performed well, yielding maximal improvements of \textgreater 0.9 DSC in accuracy, which is  particularly observed in presence of high error rates (see Fig. \ref{fig:results_mean}). Its application allows to obtain high quality (DSC \textgreater 0.9) segmentations at the maximum tested ERR of 50\%. At lower ERR, small decreases in accuracy are found. These are associated with a decrease in AR due to outlier removal. This effect can be seen when no annotation errors were present in the atlas prior to outlier correction (\emph{i.e.} ERR = 0\%). Fig. \ref{fig:results_mean} nicely illustrates the existence of an upper accuracy bound (\emph{i.e.} where ERR = 0\%) and the comparable performance at higher ERR after correction. 



\subsection*{Conclusions}
We tested forms of weak annotations to be used in medical image segmentation problems and examined the effects of expertise and frequency of annotations in their impact on accuracy outcome. Resulting segmentation accuracy was comparable to state-of-the-art performance of a fully-supervised segmentation method and the proposed outlier correction using a \emph{weakly labelled atlas} was able to largely improve the accuracy outcome for all examined types of weak annotations. The robust performance of this approach suggests that weak annotations from non-expert crowd populations could be used obtain accurate liver segmentations and the general approach can be readily adapted to other organ segmentation problems.


\section*{Acknowledgements}
We gratefully acknowledge the support of NVIDIA Corporation with the donation of a Tesla K40 GPU used for this research. This work was supported by Wellcome Trust and EPSRC IEH award [102431] for the iFIND project and the Developing Human Connectome Project, which  is funded through a Synergy Grant by the European Research Council (ERC) under the European Union's Seventh Framework Programme (FP/2007-2013) / ERC Grant Agreement number 319456.

\bibliographystyle{model2-names}
\bibliography{refs}

\begin{thebibliography}{32}
\expandafter\ifx\csname natexlab\endcsname\relax\def\natexlab#1{#1}\fi
\providecommand{\url}[1]{\texttt{#1}}
\providecommand{\href}[2]{#2}
\providecommand{\path}[1]{#1}
\providecommand{\DOIprefix}{doi:}
\providecommand{\ArXivprefix}{arXiv:}
\providecommand{\URLprefix}{URL: }
\providecommand{\Pubmedprefix}{pmid:}
\providecommand{\doi}[1]{\href{http://dx.doi.org/#1}{\path{#1}}}
\providecommand{\Pubmed}[1]{\href{pmid:#1}{\path{#1}}}
\providecommand{\bibinfo}[2]{#2}
\ifx\xfnm\relax \def\xfnm[#1]{\unskip,\space#1}\fi
\bibitem[{Albarqouni et~al.(2016)Albarqouni, Baur, Achilles, Belagiannis,
  Demirci and Navab}]{albarqouni2016aggnet}
\bibinfo{author}{Albarqouni, S.}, \bibinfo{author}{Baur, C.},
  \bibinfo{author}{Achilles, F.}, \bibinfo{author}{Belagiannis, V.},
  \bibinfo{author}{Demirci, S.}, \bibinfo{author}{Navab, N.},
  \bibinfo{year}{2016}.
\newblock \bibinfo{title}{Aggnet: Deep learning from crowds for mitosis
  detection in breast cancer histology images}.
\newblock \bibinfo{journal}{IEEE transactions on medical imaging}
  \bibinfo{volume}{35}, \bibinfo{pages}{1313--1321}.
\bibitem[{Aljabar et~al.(2009)Aljabar, Heckemann, Hammers, Hajnal and
  Rueckert}]{aljabar2009multi}
\bibinfo{author}{Aljabar, P.}, \bibinfo{author}{Heckemann, R.A.},
  \bibinfo{author}{Hammers, A.}, \bibinfo{author}{Hajnal, J.V.},
  \bibinfo{author}{Rueckert, D.}, \bibinfo{year}{2009}.
\newblock \bibinfo{title}{Multi-atlas based segmentation of brain images: atlas
  selection and its effect on accuracy}.
\newblock \bibinfo{journal}{Neuroimage} \bibinfo{volume}{46},
  \bibinfo{pages}{726--738}.
\bibitem[{Baxter et~al.(2017)Baxter, Rajchl, McLeod, Yuan and
  Peters}]{baxter2017directed}
\bibinfo{author}{Baxter, J.S.}, \bibinfo{author}{Rajchl, M.},
  \bibinfo{author}{McLeod, A.J.}, \bibinfo{author}{Yuan, J.},
  \bibinfo{author}{Peters, T.M.}, \bibinfo{year}{2017}.
\newblock \bibinfo{title}{Directed acyclic graph continuous max-flow image
  segmentation for unconstrained label orderings}.
\newblock \bibinfo{journal}{International Journal of Computer Vision} ,
  \bibinfo{pages}{1--20}.
\bibitem[{Baxter et~al.(2015)Baxter, Rajchl, Peters and
  Chen}]{baxter2015optimization}
\bibinfo{author}{Baxter, J.S.}, \bibinfo{author}{Rajchl, M.},
  \bibinfo{author}{Peters, T.M.}, \bibinfo{author}{Chen, E.C.},
  \bibinfo{year}{2015}.
\newblock \bibinfo{title}{Optimization-based interactive segmentation interface
  for multi-region problems}, in: \bibinfo{booktitle}{SPIE Medical Imaging},
  \bibinfo{organization}{International Society for Optics and Photonics}. pp.
  \bibinfo{pages}{94133T--94133T}.
\bibitem[{Boykov and Jolly(2000)}]{boykov2000interactive}
\bibinfo{author}{Boykov, Y.}, \bibinfo{author}{Jolly, M.P.},
  \bibinfo{year}{2000}.
\newblock \bibinfo{title}{Interactive organ segmentation using graph cuts}, in:
  \bibinfo{booktitle}{Medical Image Computing and Computer-Assisted
  Intervention--MICCAI 2000}. \bibinfo{publisher}{Springer Berlin Heidelberg},
  pp. \bibinfo{pages}{276--286}.
\bibitem[{Boykov and Jolly(2001)}]{boykov2001interactive}
\bibinfo{author}{Boykov, Y.Y.}, \bibinfo{author}{Jolly, M.P.},
  \bibinfo{year}{2001}.
\newblock \bibinfo{title}{Interactive graph cuts for optimal boundary \& region
  segmentation of objects in nd images}, in: \bibinfo{booktitle}{Computer
  Vision, 2001. ICCV 2001. Proceedings. Eighth IEEE International Conference
  on}, \bibinfo{organization}{IEEE}. pp. \bibinfo{pages}{105--112}.
\bibitem[{Cheplygina et~al.(2016)Cheplygina, Perez-Rovira, Kuo, Tiddens and
  de~Bruijne}]{cheplygina2016early}
\bibinfo{author}{Cheplygina, V.}, \bibinfo{author}{Perez-Rovira, A.},
  \bibinfo{author}{Kuo, W.}, \bibinfo{author}{Tiddens, H.A.},
  \bibinfo{author}{de~Bruijne, M.}, \bibinfo{year}{2016}.
\newblock \bibinfo{title}{Early experiences with crowdsourcing airway
  annotations in chest ct}, in: \bibinfo{booktitle}{International Workshop on
  Large-Scale Annotation of Biomedical Data and Expert Label Synthesis},
  \bibinfo{organization}{Springer}. pp. \bibinfo{pages}{209--218}.
\bibitem[{Estell{\'e}s-Arolas and
  Gonz{\'a}lez-Ladr{\'o}n-De-Guevara(2012)}]{estelles2012towards}
\bibinfo{author}{Estell{\'e}s-Arolas, E.},
  \bibinfo{author}{Gonz{\'a}lez-Ladr{\'o}n-De-Guevara, F.},
  \bibinfo{year}{2012}.
\newblock \bibinfo{title}{Towards an integrated crowdsourcing definition}.
\newblock \bibinfo{journal}{Journal of Information science}
  \bibinfo{volume}{38}, \bibinfo{pages}{189--200}.
\bibitem[{Gurari et~al.(2015)Gurari, Theriault, Sameki, Isenberg, Pham,
  Purwada, Solski, Walker, Zhang, Wong et~al.}]{gurari2015collect}
\bibinfo{author}{Gurari, D.}, \bibinfo{author}{Theriault, D.},
  \bibinfo{author}{Sameki, M.}, \bibinfo{author}{Isenberg, B.},
  \bibinfo{author}{Pham, T.A.}, \bibinfo{author}{Purwada, A.},
  \bibinfo{author}{Solski, P.}, \bibinfo{author}{Walker, M.},
  \bibinfo{author}{Zhang, C.}, \bibinfo{author}{Wong, J.Y.}, et~al.,
  \bibinfo{year}{2015}.
\newblock \bibinfo{title}{How to collect segmentations for biomedical images? a
  benchmark evaluating the performance of experts, crowdsourced non-experts,
  and algorithms}, in: \bibinfo{booktitle}{2015 IEEE Winter Conference on
  Applications of Computer Vision}, \bibinfo{organization}{IEEE}. pp.
  \bibinfo{pages}{1169--1176}.
\bibitem[{Haehn et~al.(2014)Haehn, Beyer, Pfister, Knowles-Barley, Kasthuri,
  Lichtman and Roberts}]{haehn2014design}
\bibinfo{author}{Haehn, D.}, \bibinfo{author}{Beyer, J.},
  \bibinfo{author}{Pfister, H.}, \bibinfo{author}{Knowles-Barley, S.},
  \bibinfo{author}{Kasthuri, N.}, \bibinfo{author}{Lichtman, J.},
  \bibinfo{author}{Roberts, M.}, \bibinfo{year}{2014}.
\newblock \bibinfo{title}{Design and evaluation of interactive proofreading
  tools for connectomics}.
\newblock \bibinfo{journal}{Computer Graphics, IEEE Transactions on} .
\bibitem[{Koch et~al.(2017)Koch, Rajchl, Bai, Baumgartner, Tong,
  Passerat-Palmbach, Aljabar and Rueckert}]{koch2017multi}
\bibinfo{author}{Koch, L.M.}, \bibinfo{author}{Rajchl, M.},
  \bibinfo{author}{Bai, W.}, \bibinfo{author}{Baumgartner, C.F.},
  \bibinfo{author}{Tong, T.}, \bibinfo{author}{Passerat-Palmbach, J.},
  \bibinfo{author}{Aljabar, P.}, \bibinfo{author}{Rueckert, D.},
  \bibinfo{year}{2017}.
\newblock \bibinfo{title}{Multi-atlas segmentation using partially annotated
  data: Methods and annotation strategies}.
\newblock \bibinfo{journal}{IEEE Transactions on Pattern Recognition and
  Machine Intelligence} .
\bibitem[{Langerak et~al.(2010)Langerak, van~der Heide, Kotte, Viergever, van
  Vulpen and Pluim}]{langerak2010label}
\bibinfo{author}{Langerak, T.R.}, \bibinfo{author}{van~der Heide, U.A.},
  \bibinfo{author}{Kotte, A.N.}, \bibinfo{author}{Viergever, M.A.},
  \bibinfo{author}{van Vulpen, M.}, \bibinfo{author}{Pluim, J.P.},
  \bibinfo{year}{2010}.
\newblock \bibinfo{title}{Label fusion in atlas-based segmentation using a
  selective and iterative method for performance level estimation (simple)}.
\newblock \bibinfo{journal}{Medical Imaging, IEEE Transactions on}
  \bibinfo{volume}{29}, \bibinfo{pages}{2000--2008}.
\bibitem[{Lin et~al.(2014)Lin, Maire, Belongie, Hays, Perona, Ramanan,
  Doll{\'a}r and Zitnick}]{lin2014microsoft}
\bibinfo{author}{Lin, T.Y.}, \bibinfo{author}{Maire, M.},
  \bibinfo{author}{Belongie, S.}, \bibinfo{author}{Hays, J.},
  \bibinfo{author}{Perona, P.}, \bibinfo{author}{Ramanan, D.},
  \bibinfo{author}{Doll{\'a}r, P.}, \bibinfo{author}{Zitnick, C.L.},
  \bibinfo{year}{2014}.
\newblock \bibinfo{title}{Microsoft coco: Common objects in context}, in:
  \bibinfo{booktitle}{Computer Vision--ECCV 2014}.
  \bibinfo{publisher}{Springer}, pp. \bibinfo{pages}{740--755}.
\bibitem[{Maier-Hein et~al.(2014)Maier-Hein, Mersmann, Kondermann, Bodenstedt,
  Sanchez, Stock, Kenngott, Eisenmann and Speidel}]{maier2014can}
\bibinfo{author}{Maier-Hein, L.}, \bibinfo{author}{Mersmann, S.},
  \bibinfo{author}{Kondermann, D.}, \bibinfo{author}{Bodenstedt, S.},
  \bibinfo{author}{Sanchez, A.}, \bibinfo{author}{Stock, C.},
  \bibinfo{author}{Kenngott, H.G.}, \bibinfo{author}{Eisenmann, M.},
  \bibinfo{author}{Speidel, S.}, \bibinfo{year}{2014}.
\newblock \bibinfo{title}{Can masses of non-experts train highly accurate image
  classifiers?}, in: \bibinfo{booktitle}{Medical Image Computing and
  Computer-Assisted Intervention--MICCAI 2014}. \bibinfo{publisher}{Springer
  International Publishing}, pp. \bibinfo{pages}{438--445}.
\bibitem[{Mavandadi et~al.(2012)Mavandadi, Feng, Yu, Dimitrov, Yu and
  Ozcan}]{mavandadi2012biogames}
\bibinfo{author}{Mavandadi, S.}, \bibinfo{author}{Feng, S.},
  \bibinfo{author}{Yu, F.}, \bibinfo{author}{Dimitrov, S.},
  \bibinfo{author}{Yu, R.}, \bibinfo{author}{Ozcan, A.}, \bibinfo{year}{2012}.
\newblock \bibinfo{title}{Biogames: A platform for crowd-sourced biomedical
  image analysis and telediagnosis}.
\newblock \bibinfo{journal}{GAMES FOR HEALTH: Research, Development, and
  Clinical Applications} \bibinfo{volume}{1}, \bibinfo{pages}{373--376}.
\bibitem[{McKenna et~al.(2012)McKenna, Wang, Nguyen, Burns, Petrick and
  Summers}]{mckenna2012strategies}
\bibinfo{author}{McKenna, M.T.}, \bibinfo{author}{Wang, S.},
  \bibinfo{author}{Nguyen, T.B.}, \bibinfo{author}{Burns, J.E.},
  \bibinfo{author}{Petrick, N.}, \bibinfo{author}{Summers, R.M.},
  \bibinfo{year}{2012}.
\newblock \bibinfo{title}{Strategies for improved interpretation of
  computer-aided detections for ct colonography utilizing distributed human
  intelligence}.
\newblock \bibinfo{journal}{Medical image analysis} \bibinfo{volume}{16},
  \bibinfo{pages}{1280--1292}.
\bibitem[{Modat et~al.(2010)Modat, Ridgway, Taylor, Lehmann, Barnes, Hawkes,
  Fox and Ourselin}]{modat2010fast}
\bibinfo{author}{Modat, M.}, \bibinfo{author}{Ridgway, G.R.},
  \bibinfo{author}{Taylor, Z.A.}, \bibinfo{author}{Lehmann, M.},
  \bibinfo{author}{Barnes, J.}, \bibinfo{author}{Hawkes, D.J.},
  \bibinfo{author}{Fox, N.C.}, \bibinfo{author}{Ourselin, S.},
  \bibinfo{year}{2010}.
\newblock \bibinfo{title}{Fast free-form deformation using graphics processing
  units}.
\newblock \bibinfo{journal}{Computer methods and programs in biomedicine}
  \bibinfo{volume}{98}, \bibinfo{pages}{278--284}.
\bibitem[{Nodine et~al.(1999)Nodine, Kundel, Mello-Thoms, Weinstein, Orel,
  Sullivan and Conant}]{nodine1999experience}
\bibinfo{author}{Nodine, C.F.}, \bibinfo{author}{Kundel, H.L.},
  \bibinfo{author}{Mello-Thoms, C.}, \bibinfo{author}{Weinstein, S.P.},
  \bibinfo{author}{Orel, S.G.}, \bibinfo{author}{Sullivan, D.C.},
  \bibinfo{author}{Conant, E.F.}, \bibinfo{year}{1999}.
\newblock \bibinfo{title}{How experience and training influence mammography
  expertise}.
\newblock \bibinfo{journal}{Academic radiology} \bibinfo{volume}{6},
  \bibinfo{pages}{575--585}.
\bibitem[{Nodine and Mello-Thoms(2000)}]{nodine2000nature}
\bibinfo{author}{Nodine, C.F.}, \bibinfo{author}{Mello-Thoms, C.},
  \bibinfo{year}{2000}.
\newblock \bibinfo{title}{The nature of expertise in radiology}.
\newblock \bibinfo{journal}{Handbook of Medical Imaging. SPIE} .
\bibitem[{Olabarriaga and Smeulders(2001)}]{olabarriaga2001interaction}
\bibinfo{author}{Olabarriaga, S.D.}, \bibinfo{author}{Smeulders, A.W.},
  \bibinfo{year}{2001}.
\newblock \bibinfo{title}{Interaction in the segmentation of medical images: A
  survey}.
\newblock \bibinfo{journal}{Medical image analysis} \bibinfo{volume}{5},
  \bibinfo{pages}{127--142}.
\bibitem[{Papandreou et~al.(2015)Papandreou, Chen, Murphy and
  Yuille}]{papandreou2015weakly}
\bibinfo{author}{Papandreou, G.}, \bibinfo{author}{Chen, L.C.},
  \bibinfo{author}{Murphy, K.}, \bibinfo{author}{Yuille, A.L.},
  \bibinfo{year}{2015}.
\newblock \bibinfo{title}{Weakly-and semi-supervised learning of a dcnn for
  semantic image segmentation}.
\newblock \bibinfo{journal}{arXiv preprint arXiv:1502.02734} .
\bibitem[{Pitiot et~al.(2004)Pitiot, Delingette, Thompson and
  Ayache}]{pitiot2004expert}
\bibinfo{author}{Pitiot, A.}, \bibinfo{author}{Delingette, H.},
  \bibinfo{author}{Thompson, P.M.}, \bibinfo{author}{Ayache, N.},
  \bibinfo{year}{2004}.
\newblock \bibinfo{title}{Expert knowledge-guided segmentation system for brain
  mri}.
\newblock \bibinfo{journal}{NeuroImage} \bibinfo{volume}{23},
  \bibinfo{pages}{S85--S96}.
\bibitem[{Rajchl et~al.(2016a)Rajchl, Baxter, McLeod, Yuan, Qiu, Peters and
  Khan}]{rajchl2016hierarchical}
\bibinfo{author}{Rajchl, M.}, \bibinfo{author}{Baxter, J.S.},
  \bibinfo{author}{McLeod, A.J.}, \bibinfo{author}{Yuan, J.},
  \bibinfo{author}{Qiu, W.}, \bibinfo{author}{Peters, T.M.},
  \bibinfo{author}{Khan, A.R.}, \bibinfo{year}{2016}a.
\newblock \bibinfo{title}{Hierarchical max-flow segmentation framework for
  multi-atlas segmentation with kohonen self-organizing map based gaussian
  mixture modeling}.
\newblock \bibinfo{journal}{Medical image analysis} \bibinfo{volume}{27},
  \bibinfo{pages}{45--56}.
\bibitem[{Rajchl et~al.(2017)Rajchl, Lee, Oktay, Kamnitsas, Passerat-Palmbach,
  Bai, Damodaram, Rutherford, Hajnal, Kainz et~al.}]{rajchl2017deepcut}
\bibinfo{author}{Rajchl, M.}, \bibinfo{author}{Lee, M.C.},
  \bibinfo{author}{Oktay, O.}, \bibinfo{author}{Kamnitsas, K.},
  \bibinfo{author}{Passerat-Palmbach, J.}, \bibinfo{author}{Bai, W.},
  \bibinfo{author}{Damodaram, M.}, \bibinfo{author}{Rutherford, M.A.},
  \bibinfo{author}{Hajnal, J.V.}, \bibinfo{author}{Kainz, B.}, et~al.,
  \bibinfo{year}{2017}.
\newblock \bibinfo{title}{Deepcut: Object segmentation from bounding box
  annotations using convolutional neural networks}.
\newblock \bibinfo{journal}{IEEE transactions on medical imaging}
  \bibinfo{volume}{36}, \bibinfo{pages}{674--683}.
\bibitem[{Rajchl et~al.(2016b)Rajchl, Lee, Schrans, Davidson,
  Passerat-Palmbach, Tarroni, Alansary, Oktay, Kainz and
  Rueckert}]{rajchl2016learning}
\bibinfo{author}{Rajchl, M.}, \bibinfo{author}{Lee, M.C.},
  \bibinfo{author}{Schrans, F.}, \bibinfo{author}{Davidson, A.},
  \bibinfo{author}{Passerat-Palmbach, J.}, \bibinfo{author}{Tarroni, G.},
  \bibinfo{author}{Alansary, A.}, \bibinfo{author}{Oktay, O.},
  \bibinfo{author}{Kainz, B.}, \bibinfo{author}{Rueckert, D.},
  \bibinfo{year}{2016}b.
\newblock \bibinfo{title}{Learning under distributed weak supervision}.
\newblock \bibinfo{journal}{arXiv preprint arXiv:1606.01100} .
\bibitem[{Rajchl et~al.(2012)Rajchl, Yuan, Ukwatta and Peters}]{rajchl2012fast}
\bibinfo{author}{Rajchl, M.}, \bibinfo{author}{Yuan, J.},
  \bibinfo{author}{Ukwatta, E.}, \bibinfo{author}{Peters, T.},
  \bibinfo{year}{2012}.
\newblock \bibinfo{title}{Fast interactive multi-region cardiac segmentation
  with linearly ordered labels}, in: \bibinfo{booktitle}{Biomedical Imaging
  (ISBI), 2012 9th IEEE International Symposium on},
  \bibinfo{organization}{IEEE Conference Publications}. pp.
  \bibinfo{pages}{1409--1412}.
\bibitem[{Rother et~al.(2004)Rother, Kolmogorov and Blake}]{rother2004grabcut}
\bibinfo{author}{Rother, C.}, \bibinfo{author}{Kolmogorov, V.},
  \bibinfo{author}{Blake, A.}, \bibinfo{year}{2004}.
\newblock \bibinfo{title}{Grabcut: Interactive foreground extraction using
  iterated graph cuts}, in: \bibinfo{booktitle}{ACM transactions on graphics
  (TOG)}, \bibinfo{organization}{ACM}. pp. \bibinfo{pages}{309--314}.
\bibitem[{Russell et~al.(2008)Russell, Torralba, Murphy and
  Freeman}]{russell2008labelme}
\bibinfo{author}{Russell, B.C.}, \bibinfo{author}{Torralba, A.},
  \bibinfo{author}{Murphy, K.P.}, \bibinfo{author}{Freeman, W.T.},
  \bibinfo{year}{2008}.
\newblock \bibinfo{title}{Labelme: a database and web-based tool for image
  annotation}.
\newblock \bibinfo{journal}{International journal of computer vision}
  \bibinfo{volume}{77}, \bibinfo{pages}{157--173}.
\bibitem[{Wolz et~al.(2012)Wolz, Chu, Misawa, Mori and
  Rueckert}]{wolz2012multi}
\bibinfo{author}{Wolz, R.}, \bibinfo{author}{Chu, C.}, \bibinfo{author}{Misawa,
  K.}, \bibinfo{author}{Mori, K.}, \bibinfo{author}{Rueckert, D.},
  \bibinfo{year}{2012}.
\newblock \bibinfo{title}{Multi-organ abdominal ct segmentation using
  hierarchically weighted subject-specific atlases}, in:
  \bibinfo{booktitle}{Medical Image Computing and Computer-Assisted
  Intervention--MICCAI 2012}. \bibinfo{publisher}{Springer Berlin Heidelberg},
  pp. \bibinfo{pages}{10--17}.
\bibitem[{Yuan et~al.(2010a)Yuan, Bae and Tai}]{yuan2010study}
\bibinfo{author}{Yuan, J.}, \bibinfo{author}{Bae, E.}, \bibinfo{author}{Tai,
  X.C.}, \bibinfo{year}{2010}a.
\newblock \bibinfo{title}{A study on continuous max-flow and min-cut
  approaches}, in: \bibinfo{booktitle}{Computer Vision and Pattern
  Recognition--CVPR 2010}, \bibinfo{organization}{IEEE}. pp.
  \bibinfo{pages}{2217--2224}.
\bibitem[{Yuan et~al.(2010b)Yuan, Bae, Tai and Boykov}]{yuan2010continuous}
\bibinfo{author}{Yuan, J.}, \bibinfo{author}{Bae, E.}, \bibinfo{author}{Tai,
  X.C.}, \bibinfo{author}{Boykov, Y.}, \bibinfo{year}{2010}b.
\newblock \bibinfo{title}{A continuous max-flow approach to potts model}, in:
  \bibinfo{booktitle}{Computer Vision--ECCV 2010}. \bibinfo{publisher}{Springer
  Berlin Heidelberg}, pp. \bibinfo{pages}{379--392}.
\bibitem[{Yushkevich et~al.(2006)Yushkevich, Piven, Hazlett, Smith, Ho, Gee and
  Gerig}]{yushkevich2006user}
\bibinfo{author}{Yushkevich, P.A.}, \bibinfo{author}{Piven, J.},
  \bibinfo{author}{Hazlett, H.C.}, \bibinfo{author}{Smith, R.G.},
  \bibinfo{author}{Ho, S.}, \bibinfo{author}{Gee, J.C.},
  \bibinfo{author}{Gerig, G.}, \bibinfo{year}{2006}.
\newblock \bibinfo{title}{User-guided 3d active contour segmentation of
  anatomical structures: significantly improved efficiency and reliability}.
\newblock \bibinfo{journal}{Neuroimage} \bibinfo{volume}{31},
  \bibinfo{pages}{1116--1128}.

\end{thebibliography}







\end{document}